\documentclass{article}

\usepackage{graphicx}
\usepackage[numbers]{natbib}

\usepackage[preprint]{neurips_2024}

\usepackage[utf8]{inputenc} 
\usepackage[T1]{fontenc}    
\usepackage{hyperref}       
\usepackage{url}            
\usepackage{booktabs}       
\usepackage{amsfonts}       
\usepackage{nicefrac}       
\usepackage{microtype}      
\usepackage{xcolor}         
\usepackage{titlesec}
\usepackage{array} 

\titlespacing*{\paragraph}{0pt}{0.3ex plus 0.1ex minus 0.2ex}{0.5em}
\titleformat{\paragraph}[runin]{\normalfont\normalsize\bfseries}{\theparagraph}{0.3em}{}[]

\title{CLASP: Contrastive Language-Speech Pretraining \\for Multilingual Multimodal Information Retrieval}

\author{%
  \textbf{Mohammad Mahdi Abootorabi} {\normalfont and} \textbf{Ehsaneddin Asgari} \\
  Qatar Computing Research Institute (QCRI) \\
  \texttt{mahdi.abootorabi2@gmail.com} \\
  \texttt{easgari@hbku.edu.qa} \\
}

\begin{document}

\maketitle

\begin{abstract}
This study introduces CLASP (Contrastive Language-Speech Pretraining), a multilingual, multimodal representation tailored for audio-text information retrieval. CLASP leverages the synergy between spoken content and textual data. During training, we utilize our newly introduced speech-text dataset, which encompasses 15 diverse categories ranging from fiction to religion. CLASP's audio component integrates audio spectrograms with a pre-trained self-supervised speech model, while its language encoding counterpart employs a sentence encoder pre-trained on over 100 languages. This unified lightweight model bridges the gap between various modalities and languages, enhancing its effectiveness in handling and retrieving multilingual and multimodal data. Our evaluations across multiple languages demonstrate that CLASP establishes new benchmarks in HITS@1, MRR, and meanR metrics, outperforming traditional ASR-based retrieval methods that rely on transcribing speech into text for subsequent text retrieval, especially in specific scenarios.
\end{abstract}

\section{Introduction}
Multimodal machine learning is a branch of artificial intelligence that aims to integrate and model multiple forms of data, such as text, images, and audio, to capture their common meaning and enable cross-modal applications \cite{elliott-etal-2016-multimodal,morency-baltrusaitis-2017-multimodal,hessel-lee-2020-multimodal,huang-etal-2021-multilingual,yang-etal-2023-shot-joint}. This field has recently garnered considerable attention in the machine learning community. The advent of multimodal foundation models has significantly changed the landscape of machine learning tasks, particularly in natural language processing (NLP) and information retrieval \cite{chen-etal-2022-murag,wang2023largescale}. Despite the field’s primary focus on text and vision data, the speech modality often remains underexplored. This oversight occurs even though speech content is easier to generate for human beings and plays a crucial role in human cognition and interaction. Speech and text, which are two distinctive modalities of human language, fundamentally share substantial semantic similarities. These complementary forms of communication raise an intriguing question: how can they be represented to encapsulate their inherent shared meaning effectively? Such a representation is crucial for facilitating cross-modal applications including, but not limited to, speech-to-text translation~\cite{cattoni2021must}, speech/text retrieval~\cite{karakos-etal-2020-reformulating}, and speech summarization~\cite{manakul2020abstractive}. 

The current progress in self-supervised learning (SSL) has permeated the field of speech processing, addressing a broad array of challenges \cite{10.1109/TASLP.2021.3122291,mohamed2022self,liu2022audio}. Speech SSL techniques are broadly categorized into masked reconstruction \cite{10.1109/TASLP.2021.3122291}, contrastive learning \cite{ye-etal-2022-cross}, classification \cite{banerjee2020comparison}, multi-task learning \cite{cai2021speech}, and knowledge distillation \cite{ni2023adaptive}. Beyond unimodal SSL approaches, the integration of multimodal data is proposed further to enhance the performance of speech models in different tasks, e.g., speaker identification \cite{xiong2022look} and emotion detection \cite{liu2022multi}. One of the main challenges in multimodal speech is aligning the latent spaces of audio and text in a common embedding space. This challenge, similar to those faced by language-vision models, e.g., CLIP \cite{DBLP:conf/icml/RadfordKHRGASAM21} and Flamingo \cite{alayrac2022flamingo}, has been particularly addressed in recent literature and is the focus of this work, primarily for the information retrieval.

\section{Related Works}
Recent works have focused on the cross-modal embedding of text and speech by working on alignment and joint representation learning in Automatic Speech Recognition (ASR) and language models using fusion networks. \cite{9747760} presents a strategy involving an embedding aligner and modality switch training for bridging the mismatch between speech and text encoders, achieving notable reductions in word error rate (WER). SAMU-XLSR \cite{9834099}, focuses on learning multimodal multilingual speech embeddings at the sentence level. Maestro model \cite{chen22r_interspeech} is introduced to learn unified representations of both speech and text modalities, showing considerable improvements in various downstream tasks such as ASR and Speech Translation (ST). The unsupervised Cross-Modal Alignment \cite{NEURIPS2018_1ea97de8}, offers a framework for aligning speech and text embedding spaces in an unsupervised fashion, beneficial for low or zero-resource languages. SpeechUT \cite{zhang-etal-2022-speechut} proposes a unified-modal speech-unit-text pre-training model, ensuring improved performance on ASR and ST tasks by aligning speech and text representations. In a similar vein, SpeechBERT \cite{chuang20b_interspeech} presents a jointly learned audio-and-text model for end-to-end Spoken Question Answering, emphasizing the effectiveness of the end-to-end approach in handling ASR errors. The broader perspective of multimodal machine learning, involving the integration of language, vision, and speech, is discussed in \cite{morency-baltrusaitis-2017-multimodal}. SpeechCLIP \cite{10022954} is an example of efforts to bridge speech and text through images to enhance speech models without requiring transcriptions. Acoustic Neighbor Embeddings \cite{jeon2022acousticneighborembeddings} propose a novel method for mapping speech or text sequences into a shared low-dimensional space at the word level, enabling tasks such as isolated word recognition while preserving phonetic similarity through Euclidean distances. Generative Spoken Language Modeling \cite{lakhotia-etal-2021-generative} explores unsupervised learning of linguistic and acoustic representations directly from raw audio, offering insights into generative approaches for speech understanding.

While many studies have investigated the alignment of speech-text embedding models, they often focus on specific datasets and models, which can limit their generalizability to new domains. Furthermore, much emphasis has been placed on the ASR task, leaving an evaluation gap for joint embeddings in retrieval tasks, e.g., searching textual content within audio sources, such as a lecture or movie. These challenges highlight the necessity for more robust and general speech-text embedding approaches suitable for diverse retrieval contexts. Our research addresses this gap. While SAMU-XLSR \cite{9834099} is most similar to our work, their model's unavailability necessitates establishing new baselines with our model's backbones.

\paragraph{Contributions}
In this work \textbf{(i)} we introduce CLASP (Contrastive Language-Speech Pretraining) \footnote{Models: \href{https://huggingface.co/llm-lab/CLASP}{\url{https://huggingface.co/llm-lab/CLASP}}}, a novel lightweight multilingual, multimodal representation designed for audio-text retrieval
\footnote{Code: \href{https://github.com/language-modeling-lab/CLASP}{\url{https://github.com/language-modeling-lab/CLASP}}}.
\textbf{(ii)} We introduce a diverse paired speech-text dataset (Speech Brown) in 15 categories, encompassing a wide range of topics from fiction to religion \footnote{Proposed Dataset: \href{https://huggingface.co/datasets/llm-lab/SpeechBrown}{\url{https://huggingface.co/datasets/llm-lab/SpeechBrown}}}.
\textbf{(iii)} We show that combining audio spectrograms with pre-trained self-supervised speech embeddings enhances audio encoding for retrieval applications.
\textbf{(iv)} Evaluations in multiple languages demonstrate that CLASP sets new benchmarks in HITS@1, Mean Reciprocal Rank (MRR), and Mean Rank (meanR) metrics.

\section{Approach}

\subsection{Dataset}
For the effective training of a joint speech-text model, it is essential to have a rich speech-text parallel dataset that spans a variety of contexts and domains. The dataset generation pipeline is shown in part b of Figure \ref{fig:panel}. In this study, we utilize the following key datasets:
\paragraph{(i) Common Voice V4 \cite{commonvoice:2020}}
A comprehensive crowdsourced collection by Mozilla, this dataset is a staple in speech recognition studies. It comprises approximately 55K paired sentences and their voice recordings.

\paragraph{(ii) Fleurs \cite{10023141}}
Developed by Google and primarily utilized in speech recognition studies, this dataset contains around 3.6K pairs of sentences and corresponding speech, especially in content related to translation.

\paragraph{(iii) Synthetic Speech Brown}
To enhance the linguistic diversity and domain coverage, we enriched these datasets by employing the \textit{Brown Corpus} \cite{francis1979brown,schubert-tong-2003-extracting}. 
We created an audio version of approximately 55K sentences from the Brown Corpus, covering 15 different domain categories: ‘adventure’, ‘belles lettres’, ‘editorial’, ‘fiction’, ‘government’, ‘hobbies’, ‘humor’, ‘learned’, ‘lore’, ‘mystery’, ‘news’, ‘religion’, ‘reviews’, ‘romance’, and ‘science fiction’.
These synthetic recordings were generated using the Tacotron 2 text-to-speech model \cite{8461368}. 

In aggregate, our combined datasets present over 110K sentences across diverse categories and speakers. 
Dataset statistics are provided in Table \ref{tab:datasets-table}. 
This rich collection enhances the model's linguistic capabilities, ensuring robust adaptability across distinct domains. Additional information about Speech Brown, including sentence distribution by category, can be found in Appendix~(\S\ref{app:dataset}).

\begin{table}[t]
\caption{Overview of CLASP training dataset. We introduce ``Speech Brown'', an audio version of the Brown Corpus. Listed attributes include sample count, unique tokens, and average tokens per sentence are provided for Common Voice V4, FLEURS, and Speech Brown. Speech Brown which is introduced by this study covers 15 diverse categories.\\}
\label{tab:datasets-table}
\centering
\resizebox{\columnwidth}{!}{%
\begin{tabular}{>{\centering\arraybackslash}p{1.5cm}>{\centering\arraybackslash}p{1.5cm}>{\centering\arraybackslash}p{1.3cm}>{\centering\arraybackslash}p{1.7cm}>{\centering\arraybackslash}p{1.7cm}}
\multicolumn{1}{c}{Datasets} &\multicolumn{1}{c}{No. of Sentences}  &\multicolumn{1}{c}{Unique Tokens}  &\multicolumn{1}{c}{Avg Tokens/Sentence} &\multicolumn{1}{c}{Subjects}

\\
\hline
\\
Common Voice V4         &55,689  &52,580 & 10.09 & Blogs, Books, Movies, etc.\\
\addlinespace
\hline
\addlinespace

FLEURS          &3,643 &8,527 &21.45 &Translation-related Phrases\\
\addlinespace
\hline
\addlinespace
Speech Brown (ours)           &55,173 &50,667 &19.00 & 15 Categories.\\

\addlinespace
\hline
\addlinespace
Total    &114,117 &86,601 &14.73 \\

\end{tabular}
}
\end{table}

\vspace{-0.1em}
\subsection{Model}
In this section, we present our proposed method for the shared space embedding of audio and text, leveraging a comprehensive parallel dataset of aligned speech and text that we meticulously created in a prior step. Our strategy involves encoding both modalities and aligning the resulting embeddings.
We trained two models based on different loss functions for embedding alignment:

\textbf{(i) LASP}: This model uses Huber Loss, effectively balancing sensitivity to outliers with robust handling of small errors, making it ideal for noisy or imperfect alignment data, as demonstrated in studies such as \cite{runz2020frodo}.

\textbf{(ii) CLASP}: This model employs Contrastive Loss as in \cite{DBLP:conf/icml/RadfordKHRGASAM21} to be able to learn from both positive and negative samples.

\begin{figure}[t]
\centering
  \includegraphics[width=\textwidth]{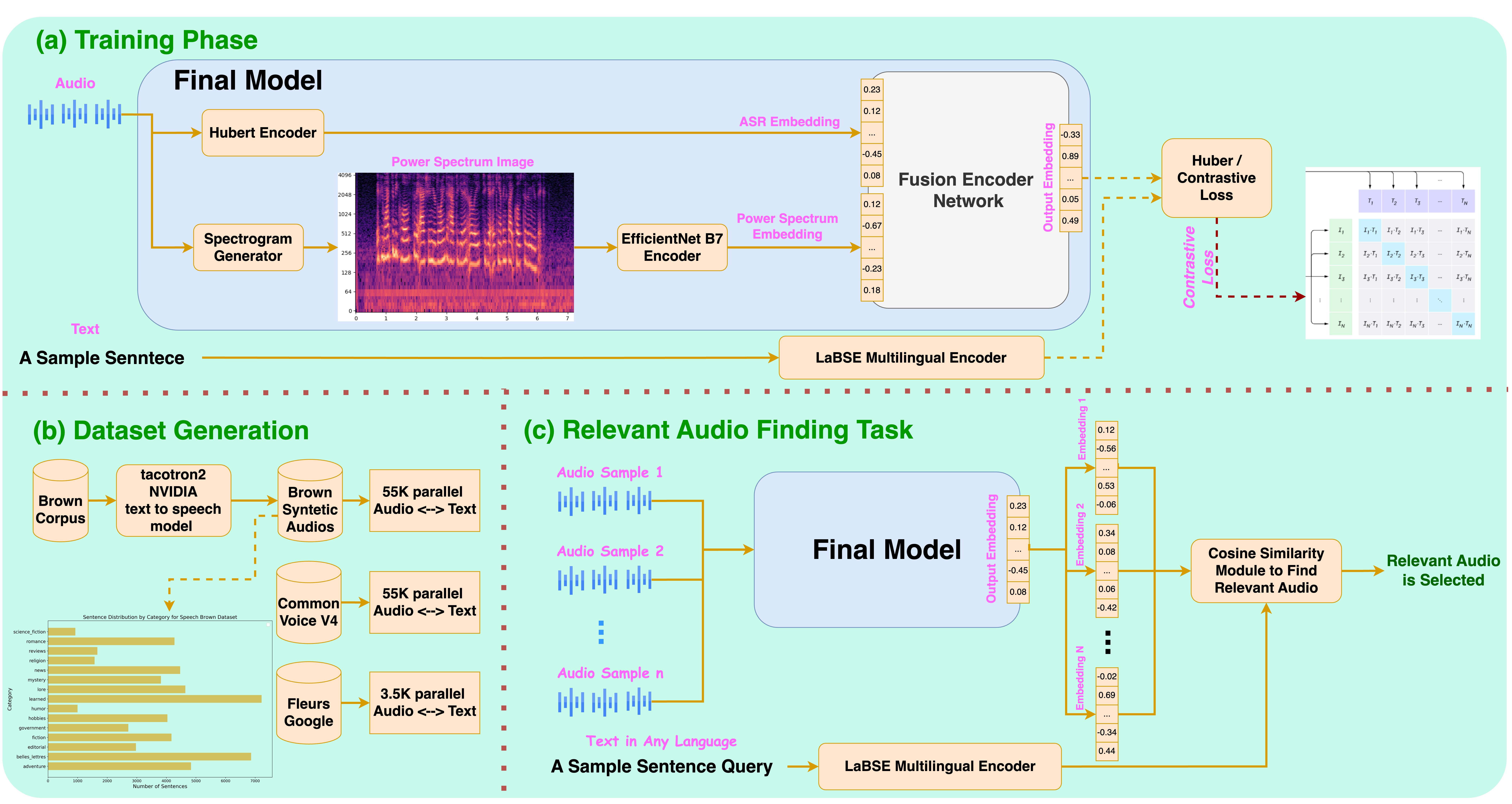}
\caption{\label{fig:panel}
Overview of the pipelines and model architecture.
\textbf{(a):} The training pipeline and architecture for CLASP, utilizing batches of speech inputs paired with their corresponding text data.
\textbf{(b):} The process of generating the final dataset for model training and evaluation.
\textbf{(c):} The inference pipeline for retrieving the exact or nearest audio from the test dataset that matches a given text query in any language.
}
\end{figure}

The training pipeline is shown in part a of Figure \ref{fig:panel}.

\subsubsection{\textbf{Speech Encoding}}
To generate the speech encoding, we employed a combination of self-supervised speech encoding and speech spectrogram embedding. We leveraged pre-trained encoders and utilized their feature extraction capabilities to obtain these encodings. For the self-supervised learning (SSL) embedding of speech, we consider two alternatives: the Wav2Vec2 model~\cite{baevski2020wav2vec} and the HuBERT model~\cite{10.1109/TASLP.2021.3122291}. To obtain an encoding representation, we applied average pooling on the outputs of these pre-trained encoders.
For the spectrogram, we initially designed a spectrogram module generator to create the power spectrum view over time. Subsequently, to encode the speech spectrogram, we employed the EfficientNet image encoder \cite{efficentnet}. The rationale behind using the speech spectrogram is that it provides a time-frequency representation of the speech signal, which can reveal important and more abstract features and characteristics of the speech that complement the self-supervised speech encodings.
After extracting these two encodings, our fusion encoder network processes them to generate the final speech embedding. We explore two distinct strategies for integrating these encodings: 
\paragraph{(i) Concatenation Strategy}
In the first strategy, each input embedding undergoes a transformation through a feed-forward network. We then concatenate the speech and spectrogram embeddings and pass them through a neural network encoder. This encoder, which consists of linear, batch normalization, dropout, and activation layers, is designed to output the final embeddings. 
\paragraph{(ii) Gating Mechanism}
After the initial transformations, the second strategy employs a gating mechanism, inspired by LSTM \cite{10.1162/neco.1997.9.8.1735} with some modifications. This mechanism controls the influence of the speech embedding and the spectrogram embedding in constructing the final encoding, determining the optimal balance of information from each modality. A softmax function then assigns weights for a weighted average, yielding the final embedding.

\begin{figure}
\centering
  \includegraphics[width=\columnwidth]{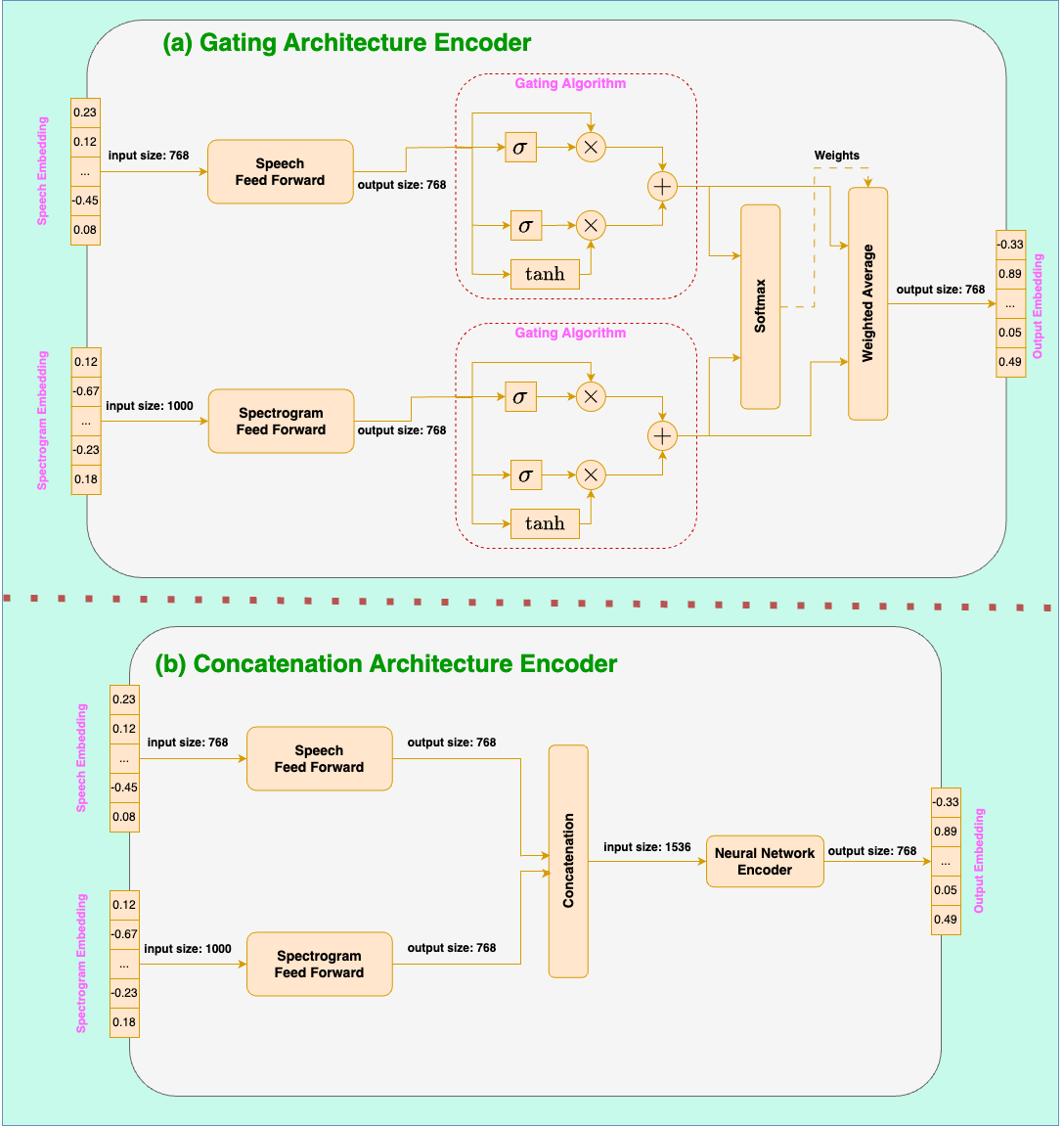}
\caption{\label{fig:encoder}
Overview of the two proposed strategies for the fusion encoder network architecture.  
\textbf{(a) Gating Mechanism}: A gating mechanism determines the contribution of self-supervised speech and spectrogram embeddings after initial transformations. 
\textbf{(b) Concatenation Strategy}: After initial transformations, the resulting self-supervised speech and spectrogram embeddings are concatenated and passed through a neural network encoder comprising linear, batch normalization, dropout, and activation layers to generate the final encoding. 
}
\end{figure}

Figure \ref{fig:encoder} shows these fusion encoder network architectures.

\subsubsection{\textbf{Text Encoding}} 
For text encoding, we adopt pre-trained multilingual language models for multilingual search within audio content. We explore two powerful multilingual backbones for text embedding: 

\textbf{(i) XLM-Roberta} \cite{conneau-etal-2020-unsupervised}, which is capable of handling multiple languages and providing robust text embeddings. 

\textbf{(ii) LaBSE} \cite{feng-etal-2022-language}, which is trained on parallel sentences encoded using the BERT encoder with a form of contrastive loss, enhancing its ability to represent multilingual text effectively. 

Both the XLM-Roberta and LaBSE models serve as powerful encoders for our text data, enabling us to achieve comprehensive and multilingual audio-text embeddings. The frozen text encoder provides a stable reference for guiding the embedding space of the speech encoding module.

\begin{table}[t!]
    \centering
    \caption{Different stages of evaluation on our test dataset for both retrieval and classification tasks. The metrics HITS@1, MRR, and meanR are used to evaluate the information retrieval experiment, while Macro-F1 assesses the performance of the classification task. For more details about each stage setup refer to the Evaluation subsection.}
    \vspace{1em}
    
    \resizebox{\textwidth}{!}{%
        \begin{tabular}{c|>{\centering\arraybackslash}p{2.0cm}>{\centering\arraybackslash}p{2.0cm}>{\centering\arraybackslash}p{2.0cm}}
            \multicolumn{4}{c}{Stage 1: Performance Comparison of Models Based on Different Modalities} \\[0.07cm]
            \hline
            \addlinespace
            Model & HITS@1 & MRR & Macro-F1 \\ \hline \addlinespace
            Spectrogram-Based Model & 0.399 & 0.610 & 0.574 \\[0.07cm]
            Self-Supervised Audio Model & 0.904 & 0.947 & 0.747 \\[0.07cm]
            \textbf{Combined Audio-Spectrogram Model} & \textbf{0.935} & \textbf{0.965} & \textbf{0.792} \\
            \addlinespace \addlinespace \addlinespace \addlinespace
            \multicolumn{4}{c}{Stage 2: Performance Comparison of Different Encoder Backbones and Fusion Encoder Architectures} \\[0.07cm]
            \hline \addlinespace 
            Model Architecture & HITS@1 & MRR & Macro-F1 \\ \hline \addlinespace
            XLM-RoBERTa + Wav2Vec2 - Gating & 0.850 & 0.912 & 0.594 \\[0.07cm]
            XLM-RoBERTa + Wav2Vec2 - Concatenation & 0.866 & 0.922 & 0.801 \\[0.07cm]
            LaBSE + Wav2Vec2 - Gating & 0.949 & 0.972 & 0.700 \\[0.07cm]
            LaBSE + Wav2Vec2 - Concatenation & 0.966 & 0.982 & 0.858 \\[0.07cm]
            XLM-RoBERTa + HuBERT - Gating & 0.993 & 0.996 & 0.835 \\[0.07cm]
            XLM-RoBERTa + HuBERT - Concatenation & 0.996 & 0.997 & 0.950 \\[0.07cm]
            LaBSE + HuBERT - Gating & 0.998 & 0.999 & 0.893 \\[0.07cm]
            \textbf{LaBSE + HuBERT - Concatenation} & \textbf{0.999} & \textbf{0.999} & \textbf{0.956} \\
        \end{tabular}
    }

    \vspace{2em}

    \resizebox{\textwidth}{!}{%
        \begin{tabular}{c|>{\centering\arraybackslash}p{2.1cm}>{\centering\arraybackslash}p{2.1cm}>{\centering\arraybackslash}p{2.1cm}>{\centering\arraybackslash}p{2.1cm}}
            \multicolumn{5}{c}{Stage 3: Performance Comparison of CLASP, LASP, and ASR-based Baselines} \\[0.07cm]
            \hline \addlinespace
            Model & HITS@1 & MRR & meanR & Macro-F1 \\ \hline \addlinespace
            LASP (ours) & 0.878 & 0.909 & 9.092 & 0.534 \\[0.07cm]
            Wav2Vec2 ASR & 0.927 & 0.939 & 38.300 & 0.678 \\[0.07cm]
            \textbf{CLASP (ours)} & \textbf{0.940} & \textbf{0.955} & \textbf{7.710} & \textbf{0.710} \\[0.07cm]
            HuBERT ASR & \textbf{0.953} & \textbf{0.963} & 17.840 & 0.679 \\
            \addlinespace \addlinespace \addlinespace
            \multicolumn{5}{c}{Stage 4: Performance of the Optimal Proposed Model in Non-English Languages} \\[0.07cm]
            \hline \addlinespace
            Language & HITS@1 & MRR & meanR & Macro-F1 \\ \hline \addlinespace
            Persian & 0.794 & 0.848 & 26.960 & 0.670 \\[0.07cm]
            German & 0.812 & 0.863 & 21.637 & 0.671 \\[0.07cm]
            French & 0.849 & 0.892 & 16.100 & 0.718 \\[0.07cm]
            Chinese & 0.822 & 0.871 & 27.125 & 0.743 \\
        \end{tabular}
    }
    \label{table_performance}
\end{table}

\subsection{Evaluation}
One primary objective of developing joint speech/text embeddings is to facilitate multilingual text search within audio content, eliminating the need for Automatic Speech Recognition (ASR). 
We assessed text-speech embeddings for multimodal retrieval and binary classification to evaluate text-speech semantic alignment. Metrics reported include MRR, HITS@1, meanR (stages 3 and 4 only) for retrieval, and Macro-F1 for classification. MRR evaluates overall ranking accuracy, while HITS@1 specifically measures the success rate of retrieving the top-ranked result. MeanR provides additional insight by quantifying the average rank of relevant results, highlighting the system’s ability to handle outliers and maintain consistent performance. Macro-F1, a robust metric for classification, equally weights all classes, making it well-suited for imbalanced datasets.

Our dataset was divided into training, development, and testing sets using an 80-10-10 split. Shared space embeddings were derived from speeches using our model, while text embeddings were obtained from queries via pre-trained text encoder. Retrieval of relevant speeches involved calculating cosine similarity. For classification, we applied a threshold of $0.5$ on cosine similarity scores to assess query relevance to speeches. The training was performed using Adam optimizer \cite{Kingma2014AdamAM} with a learning rate of $5 \times 10^{-6}$ and a batch size of 32. 
The evaluation process is illustrated in part c of Figure \ref{fig:panel}. The evaluation stages are as follows:

\paragraph{1. Various Modality Analysis} 
During this stage, we trained three models that incorporate different modalities. The first model used only the speech spectrogram embedding. The second model solely relied on self-supervised speech encoding. The third model combined both encoding methods, as proposed in the Model section. All models were trained for 50 epochs, employed a concatenation strategy for the fusion encoder neural network, evaluated with five random speech candidates per query, and optimized with Huber loss.

\paragraph{2. Effect of Backbone and Fusion Encoder Network Architecture}
During this stage, we examined all combinations of self-supervised speech encoders, text encoders, and the fusion encoder architectures mentioned earlier to evaluate their impact on performance. All models were trained with early stopping, optimized using Huber loss, and evaluated with five random speech candidates per query.

\paragraph{3. Loss Effect Analysis and ASR-Based Baselines Comparison}
At this stage, we use the optimal settings from prior stages. These included using LaBSE and HuBERT models as encoders and a concatenation strategy for the fusion encoder neural network architecture. We then trained the model using two proposed loss functions and compared them to ASR-based baselines (Wav2Vec2 and HuBERT ASR), which transcribed speeches for unimodal text retrieval. To ensure a more rigorous evaluation, all models used early stopping and were assessed using all test dataset speeches as candidates per query.

\paragraph{4. Multilingual Evaluation}
Leveraging a multilingual text encoder allows us to conduct multilingual queries within the speech vector database. The dataset was translated into French, German, Chinese, and Persian using the SMALL-100 Model translator \cite{mohammadshahi-etal-2022-small}. The optimal model from stage 3 was evaluated under the same challenging conditions.

\section{Results}
Our results for all stages are summarized in Table \ref{table_performance}. To ensure the robustness of our findings, we evaluated our models on a test dataset comprising approximately 11K samples, providing a strong statistical foundation for comparisons among models.
In stage 1, the combination of spectrogram and speech SSL model yielded the highest performance, prompting us to integrate spectrogram embeddings with the SSL model in later stages. 
In stage 2, we found that pairing the LaBSE text encoder with HuBERT embeddings and employing the concatenation strategy for the fusion encoder architecture achieved the highest retrieval scores, which we then adopted for subsequent evaluations.

Stage 3 results show that training with contrastive loss enhances the model’s ability to capture semantics. This suggests that the model not only learns from positive samples but also differentiates between negative samples, thereby leading to more robust and discriminative representations. 
Furthermore, compared to ASR-based models, CLASP performs better than Wav2Vec2 and nearly matches HuBERT, while offering significant advantages in model size (reduced by approximately 50\%) and inference speed (improved by around 10\%). Notably, CLASP was trained on approximately 90K samples (about 130 hours), considerably less than the over 60K hours used by HuBERT. Additionally, CLASP’s size is about 1.5 GB, substantially smaller than the 3.15 GB HuBERT-ASR pipeline. CLASP's lower meanR values compared to ASR-based models indicate superior handling of outliers and improved retrieval positions for relevant speech. 
Stage 4 results confirm that CLASP demonstrates exceptional and consistent performance in multilingual retrieval scenarios, as evidenced by evaluations in French, German, Chinese, and Persian.

Finally, as suggested by Figure \ref{fig:tsne}, the projection of our model’s embedding space into the multilingual text encoder space effectively establishes a shared embedding space. These results confirm CLASP's capability to generate semantic, multilingual embeddings directly from raw speech, potentially capturing information beyond ASR capabilities.

\definecolor{x}{rgb}{0.0, 0.4, 0.0} 

\begin{figure}
\begin{center}
  \includegraphics[width=\columnwidth]{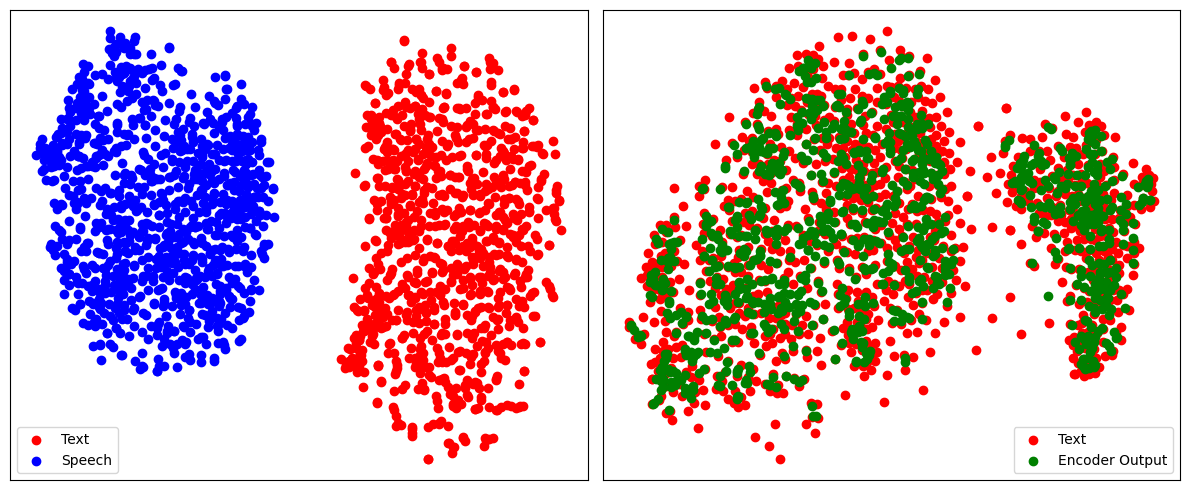}
\end{center}
\caption{\label{fig:tsne}
2-D illustration of sentence-level embeddings from different modalities, showing effective projection in the shared representation space for the test dataset. On the left, the self-supervised speech embeddings (\textcolor{blue}{blue}) and text embeddings (\textcolor{red}{red}) are depicted. On the right, the CLASP output embeddings for speech (\textcolor{x}{green}) and text embeddings (\textcolor{red}{red}) are presented. These plots were generated using the t-SNE \cite{vanDerMaaten2008} dimensionality reduction technique.
}
\end{figure}

\section{Conclusions}

In summary, we presented CLASP, a specialized multilingual, multimodal representation for audio-text retrieval that leverages the synergy between spoken content and text. CLASP integrates audio spectrograms with a self-supervised speech model and a multilingual sentence encoder, trained on our newly introduced Speech Brown dataset alongside two standard datasets, supporting nearly 100 languages. Our comprehensive evaluations demonstrate that this unified lightweight model establishes a new performance benchmark in audio-text retrieval across multiple languages without the need to transcribe speeches, as required in ASR-based pipelines. This unified model effectively bridges the gap between speech and language in information retrieval.
In the future, we aim to enhance the comprehensiveness of the CLASP embeddings. One potential avenue for this could involve incorporating speaker attributes and other non-semantic information, such as gender or emotional state, into embeddings to expand their applicability to tasks such as emotion recognition and other speech understanding challenges beyond the capabilities of ASR models.

\section{Limitations}

There are certain limitations associated with this work that have not been explored within the scope of this project. A subset of potential limitations is as follows: \textbf{(i)} The scalability of CLASP to handle very large or real-time datasets remains unclear, and the paper does not address potential challenges related to model scalability. \textbf{(ii)} CLASP relies on the availability of audio data suitable for speech encoder models, which may limit its application in scenarios where audio recordings are of low quality. \textbf{(iii)} CLASP primarily focuses on the information retrieval task, and we have not explored its potential applications in ASR or other related applications within the scope of this project.

\bibliographystyle{IEEEtran}
\bibliography{neurips_2024}

\newpage
\appendix
\section{Speech Brown Additional Details}
\label{app:dataset}
The Speech Brown Dataset comprises a substantial collection of speech and text pairs with the following characteristics:

\begin{itemize}
\item Total size: Approximately 30 GB
\item Number of samples: 55,173 pairs of speech and text
\item Maximum tokens in a sample: 48
\item Average characters per sample: 96.72
\end{itemize}

\begin{figure}
\centering
  \includegraphics[width=\columnwidth]{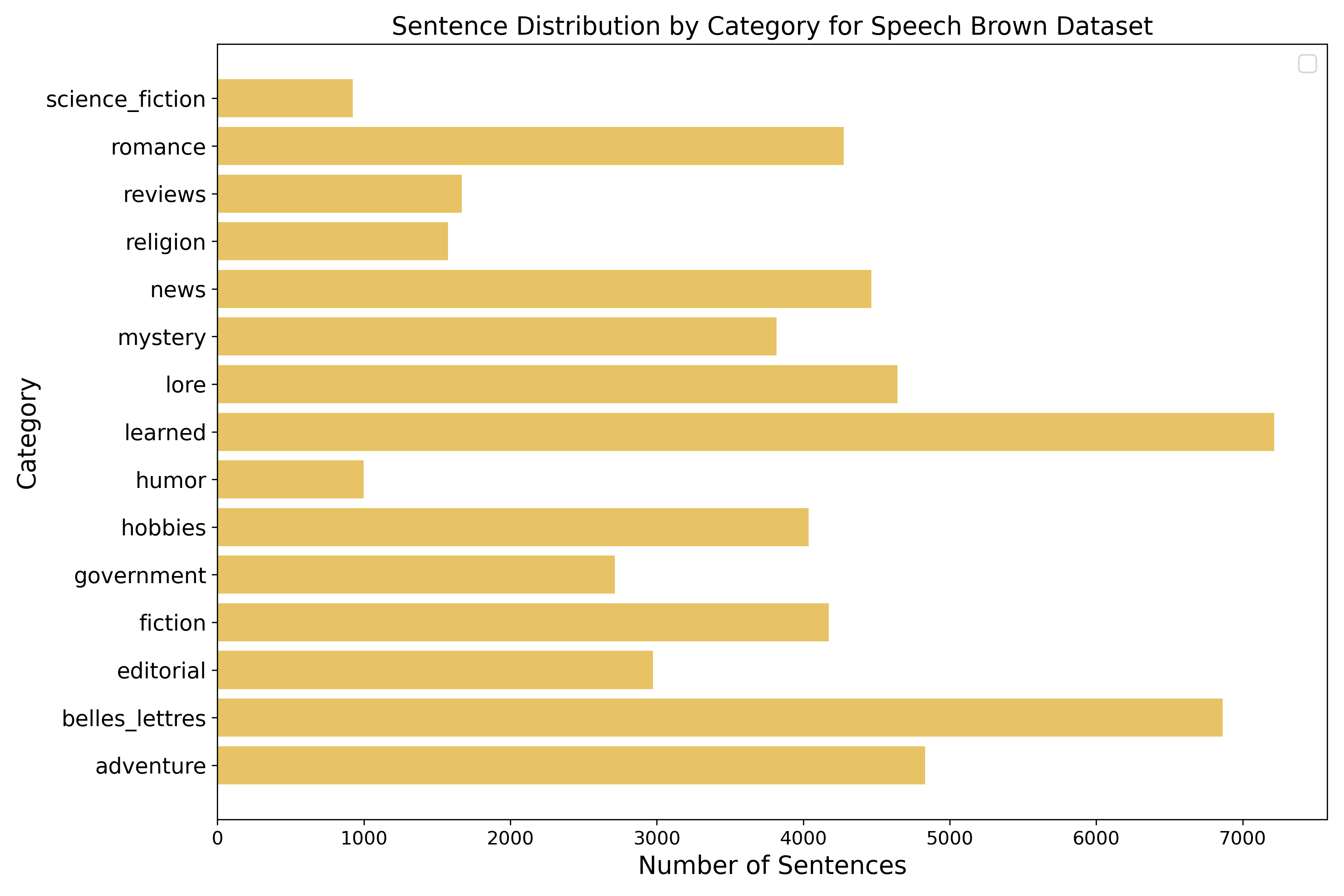}
\caption{\label{fig:brown}
Sentence Distribution by Category for Speech Brown Dataset. This horizontal bar chart illustrates the number of sentences across 15 different textual categories in the Speech Brown Dataset. 
}
\end{figure}

As shown in Figure \ref{fig:brown}, the dataset contains a diverse distribution of sentences across 15 different categories, with ‘learned’ and ‘belles\_lettres’ having the highest representation (approximately 7,000 sentences each), while categories such as ‘humor’ and ‘science\_fiction’ have considerably fewer samples (approximately 1,000 sentences each). This distribution highlights the dataset's emphasis on academic and literary content while maintaining coverage across various genres including fiction, news, and specialized domains.

The extensive vocabulary size of 50,667 unique tokens indicates the rich linguistic diversity captured in this dataset, making it particularly valuable for our research on speech processing and natural language understanding tasks.

\end{document}